\documentclass[preprint,12pt]{elsarticle}

\usepackage{graphicx}%
\usepackage{multirow}%
\usepackage{amsmath,amssymb,amsfonts}%
\usepackage{amsthm}%
\usepackage{mathrsfs}%
\usepackage[title]{appendix}%
\usepackage{xcolor}%
\usepackage{textcomp}%
\usepackage{manyfoot}%
\usepackage{booktabs}%
\usepackage{algorithm}%
\usepackage{algorithmicx}%
\usepackage{algpseudocode}%
\usepackage{listings}%
\usepackage{makecell} %
\usepackage{hyperref}
\setcitestyle{open={},close={}}   

\journal{arXiv}

\begin{document}

\begin{frontmatter}

\title{Seal2Real: Prompt Prior Learning on Diffusion Model for Unsupervised Document Seal Data Generation and Realisation}

\author[in1]{Mingfu Yan\corref{cor1}} %
\author[in1]{Jiancheng Huang\corref{cor1}}
\author[in1,in2]{Shifeng Chen\corref{cor2}}
\cortext[cor1]{Both authors contributed equally to this research. Email: mf.yan@siat.ac.cn, jc.huang@siat.ac.cn}
\cortext[cor2]{Corresponding author. Email: shifeng.chen@siat.ac.cn}
\affiliation[in1]{organization={Shenzhen Institutes of Advanced Technology, Chinese Academy of Sciences},%
            city={Shenzhen},
            postcode={518055}, 
            state={Guangdong},
            country={China}}
\affiliation[in2]{organization={Shenzhen University of Advanced Technology},%
            city={Shenzhen},
            postcode={518107}, 
            state={Guangdong},
            country={China}}

\begin{abstract}
Seal-related tasks in document processing-such as seal segmentation, authenticity verification, seal removal, and text recognition under seals-hold substantial commercial importance. However, progress in these areas has been hindered by the scarcity of labeled document seal datasets, which are essential for supervised learning. To address this limitation, we propose Seal2Real, a novel generative framework designed to synthesize large-scale labeled document seal data. As part of this work, we also present Seal-DB, a comprehensive dataset containing 20,000 labeled images to support seal-related research.
Seal2Real introduces a prompt prior learning architecture built upon a pre-trained Stable Diffusion model, effectively transferring its generative capability to the unsupervised domain of seal image synthesis. By producing highly realistic synthetic seal images, Seal2Real significantly enhances the performance of downstream seal-related tasks on real-world data. Experimental evaluations on the Seal-DB dataset demonstrate the effectiveness and practical value of the proposed framework.
The dataset is available at \href{https://github.com/liuyifan6613/DocBank-Document-Enhancement-Dataset}{https://github.com/liuyifan6613/DocBank-Document-Enhancement-Dataset}.
\end{abstract}



\begin{keyword}
Diffusion model \sep Image generation \sep Prompt learning \sep Document processing \sep Seal generation

\end{keyword}

\end{frontmatter}

\section{Introduction}\label{sec:intro}

Seals are commonly present in Chinese documents such as contracts, agreements, statements, and notices. With the ongoing digitization of documents, an increasing number are scanned into images for digital feature extraction. However, the presence of seals significantly impacts various downstream document recognition and processing tasks~\cite{Anvari2022,Bhunia2019,Atanasiu2021,Feng2022}. For instance, seals can interfere with accurate text recognition. Consequently, numerous seal-related tasks have emerged in document processing~\cite{sealgan,yang2023docdiff,lv2021photo}, including seal segmentation, authenticity discrimination, and text recognition beneath seals-tasks of considerable commercial importance.

Despite their value, several critical challenges hinder the development of these seal-related tasks:
1) \textcolor{black}{A major bottleneck is the scarcity of large-scale, open-source datasets with high fidelity and accurate annotations.} Fully supervised training for seal segmentation and text recognition beneath seals requires datasets with ground-truth annotations (e.g., segmentation masks, textual content), \textcolor{black}{while authenticity discrimination demands a diverse set of real and convincingly fake seal images.} The performance of these tasks is highly dataset-dependent. \textcolor{black}{Unfortunately, due to privacy concerns and commercial sensitivity, most existing datasets are proprietary and unpublished, posing a substantial barrier to research advancement.}
2) Authenticity discrimination specifically relies on the availability of highly realistic fake seal images for effective classifier training, \textcolor{black}{yet generating such convincing fakes is non-trivial.}
3) Tasks such as seal segmentation, text recognition under seals, and seal removal involve costly real-data collection and manual annotation \textcolor{black}{if pursued directly on real documents at scale.}

\begin{figure}[t] 
    \centering
    \includegraphics[width=1\linewidth]{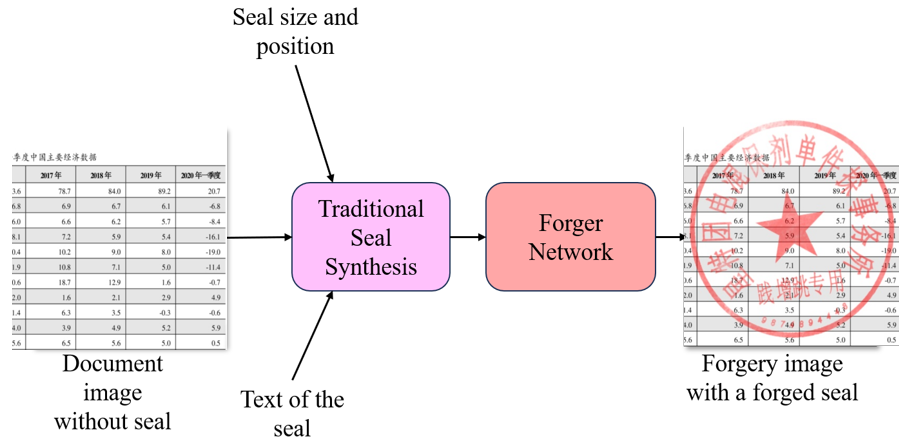} 
    \caption{Illustration of our task. We need to generate a realistic seal on a document image for making a dataset of labeled seal images.}
    \label{task}
\end{figure}

To address these limitations, synthesizing large-scale seal image datasets has become essential. 
Similar synthetic data generation and AI-augmented pipelines have demonstrated significant potential in addressing data scarcity within related fields, such as 3D content creation and immersive AR interaction~\cite{huang_ai-augmented_2025,202508.0462,huang2024ar}.
However, a critical requirement is that the synthetic data must be sufficiently realistic to support downstream applications by closely mimicking the appearance of real seals in scanned documents. In summary, a crucial upstream task is the generation of seal images. Given a document image $I_d$ without a seal, the objective is to generate a forged version $I_f$ with a synthetic seal, where the position, size, and text content of the seal in $I_f$ can be manually customized, as illustrated in Fig.~\ref{task}. This approach enables the automatic generation of corresponding labels.

Traditional approaches to seal synthesis~\cite{wen2025highly,wang2025diffusion,huangsuwan2025feddrip,alimisis2025advances,
sauer2024fast,
tang2024diffuscene,
pang2024clavaddpm,
hong2024massive} typically rely on geometric correction, template matching, and color distribution~\cite{sealgan}. However, these methods often suffer from limited robustness and require complex parameter tuning, making it difficult to generate visually convincing forged seals. \textcolor{black}{Specifically, they often fail to capture the subtle textures, ink bleed variations, paper interactions, and degradation artifacts common in real-world scanned or photographed seals.} Their effectiveness in achieving high-fidelity seal synthesis remains constrained. With the advancement of deep learning in computer vision~\cite{10268075}, modern generative models-such as GANs~\citep{xia2021gan,gal2021stylegan,kang2021gan,de2023review} and Diffusion Models~\citep{dhariwal2021diffusion,song2019generative,song2020denoising,huang2023wavedm,chen2023fec,huang2023bootstrap,huang2023kv}-offer more powerful alternatives for high-fidelity image generation. These architectures have recently pushed the boundaries of complex visual reasoning in domains like spatio-temporal autonomous driving~\cite{zeng2025FSDrive} and vision-language navigation~\cite{zeng2025janusvln}.

\textcolor{black}{However, simply applying these powerful pre-trained generative models directly to seal synthesis presents challenges. They may lack the domain-specific knowledge to generate seals with the correct structural properties and appearance nuances found in official documents. Furthermore, fine-tuning them effectively often requires large amounts of labeled domain-specific data, which is precisely what is missing for document seals. There is a clear need for a method that can adapt the strong generative capabilities of models like Stable Diffusion to the specific, unsupervised domain of realistic seal generation using limited available data.}

To bridge this gap, in our approach, we integrate Case-Based Reasoning (CBR) principles with a generative model to improve seal image synthesis. Here, each case comprises a pair: an unlabeled real seal image and a traditionally synthesized seal image, collectively forming a foundational sample library. These cases play a pivotal role across two stages of our framework. In the first stage, \textcolor{black}{they are used to guide the fine-tuning of the Stable Diffusion (SD) model through a novel prompt learning mechanism, allowing it to capture the nuanced details and stylistic variations of authentic seal designs by learning distinct representations for 'real' and 'fake' prompts based on the provided cases.} In the second stage, these same cases and the learned prompts are employed to train the forger network, guiding it to produce synthetic seals that closely resemble real ones in appearance and structure.

\textcolor{black}{By systematically leveraging these learned priors derived from comparing real and synthetic examples (inspired by CBR), our method enhances the generative model’s ability to synthesize realistic and diverse seal images in an unsupervised manner.} This integration significantly improves the authenticity and accuracy of generated seals, thereby strengthening the effectiveness and reliability of downstream document processing tasks. Overall, this demonstrates the practical value of applying prompt prior learning on diffusion models to augment machine learning-based document analysis systems.

In this paper, we propose Seal2Real, an unsupervised method for seal image generation and realization, leveraging a pretrained Stable Diffusion Model~\cite{rombach2022high}, along with a large-scale dataset, Seal-DB, comprising both real and synthetic seal images. Our primary contributions are as follows:

(1) We introduce a prompt prior learning architecture built upon a pretrained Stable Diffusion Model, which jointly optimizes the prompt embeddings and the model \textcolor{black}{using paired synthetic and unpaired real seal examples.} This enables effective and controllable seal image generation \textcolor{black}{that captures realistic seal characteristics.}

(2) We design a seal forger network that enhances the realism of synthetic seals, \textcolor{black}{guided by the learned priors,} allowing them to closely mimic real ones. This not only augments the dataset but also empowers downstream tasks-such as seal authenticity discrimination-to effectively identify forged seals.

(3) We construct Seal-DB, a comprehensive dataset generated via Seal2Real, that includes paired realistic fake seals with automatic labels and unpaired real seals. Experimental results demonstrate that our approach achieves strong performance in both seal generation and realization tasks, \textcolor{black}{leading to significant improvements in downstream applications compared to traditional synthesis.} \textcolor{black}{Our code is publicly available at https://github.com/Seal2Real/\\Seal2Real.}

\section{Method}
\label{sec:method}

\subsection{Task and Overview}

\begin{figure}[t]
\begin{center}
\includegraphics[width=0.95\linewidth]{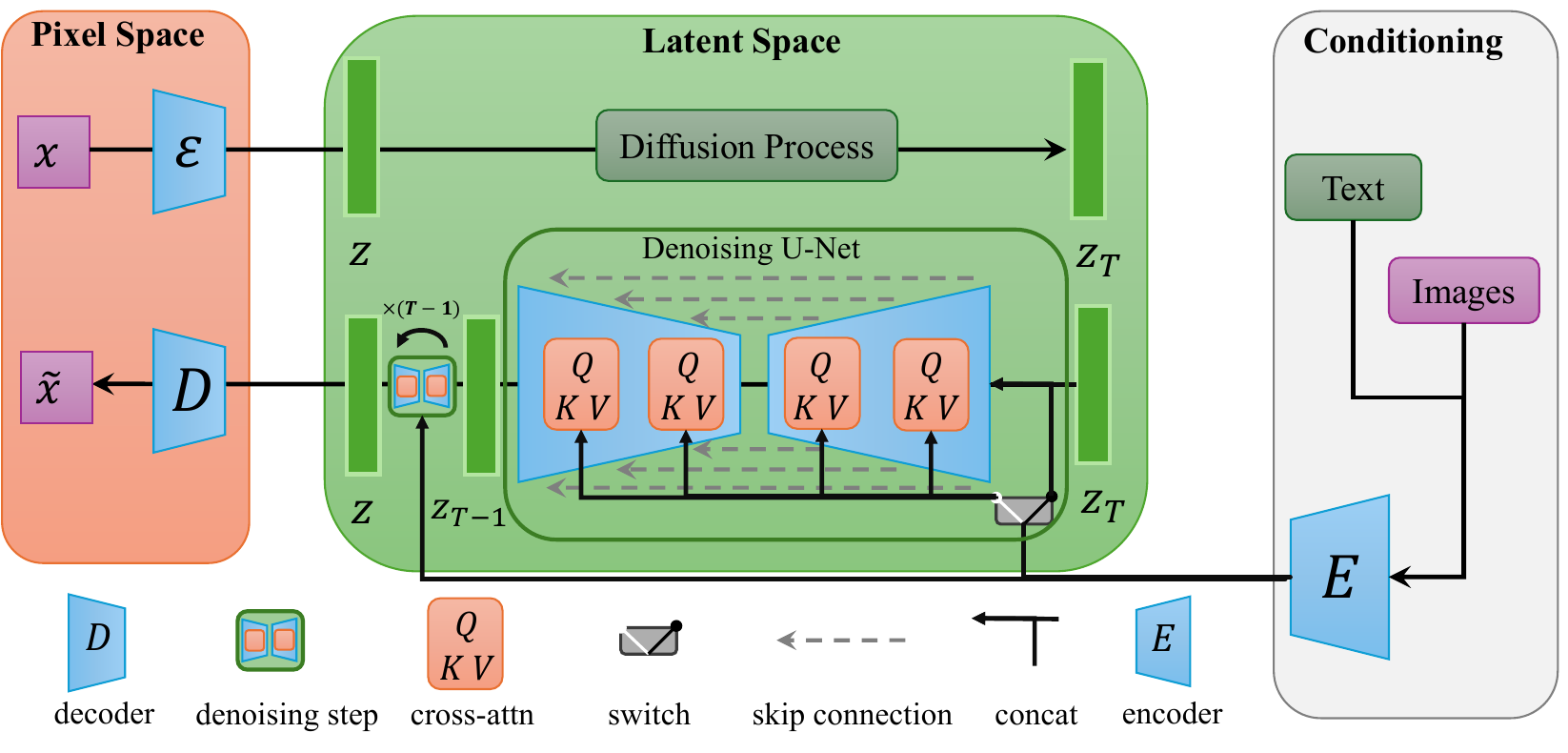}
    \end{center}
\caption{\textcolor{black}{Overview of the pretrained Stable Diffusion Model architecture employed, illustrating the key components including the Denoising U-Net. Input images $x$ are encoded ($ \mathcal{E}$) to latent space $z$. Noise is added via the diffusion process. The Denoising U-Net iteratively removes noise, conditioned on external inputs (Text/Images via Encoder $E$) using cross-attention (QKV) blocks and featuring skip connections. The final latent is decoded ($D$) back to pixel space $\tilde{x}$. This diagram provides the requested architectural information for the U-Net component.}}
\label{fig:unet}
\end{figure}

\begin{figure}[t]
\begin{center}
\includegraphics[width=0.63\linewidth]{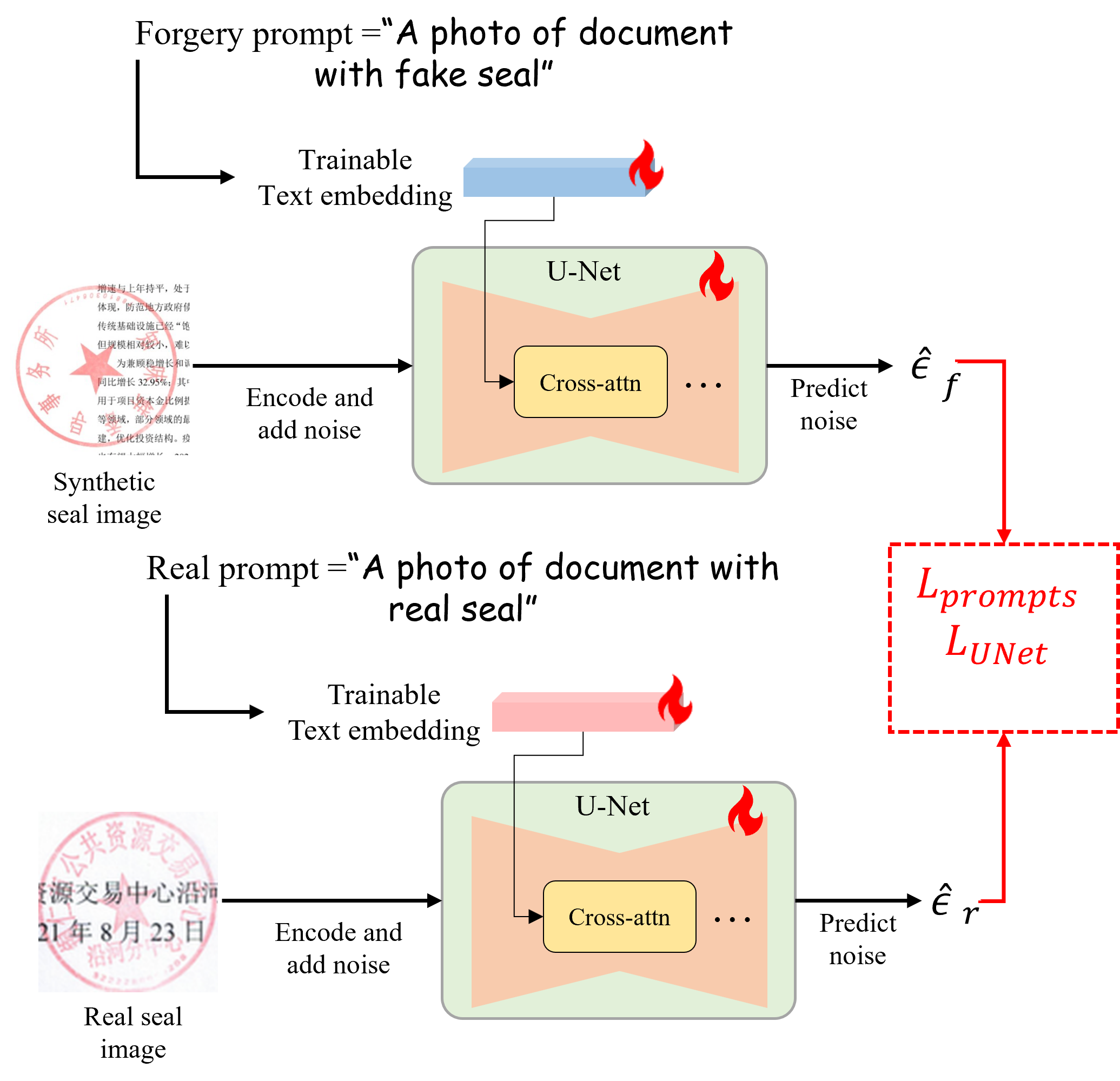}
    \end{center}
\caption{The first stage of our learning framework, termed the \textbf{Prompt Learning Stage}, focuses on learning distributions of real and fake seal images. In this stage, a diffusion loss is used to optimize both real and forgery prompts while fine-tuning the U-Net component of the Stable Diffusion (SD) model. The fine-tuned SD model can generate realistic seal images using real text prompts and synthetic seal images using forged text prompts. However, due to inherent randomness in diffusion-based generation, the output content is non-deterministic. In contrast, our proposed forger network is capable of generating forgeries with explicitly specified content. The outputs of the fine-tuned SD model are further used to train the forger in the second stage through a weakly supervised approach.}
\label{fig:prior_learning}
\end{figure}

Given a document image $I_d$ without a seal, we aim to generate a forgery image $I_f$ with a synthetic seal, where the position, size, and text of the seal in $I_f$ can be manually customized, as shown in Fig.~\ref{task}. This task involves two key challenges: 1) how to customize the position, size, and text of the seal, and 2) how to render the forged seal so realistically on the original document that it can deceive the human eye.

To address the first challenge, we adopt a traditional seal synthesis method based on the work in \cite{yang2023docdiff}, as described in Sec.~\ref{sec:intro}. This method utilizes tools such as Pillow to generate seals with customizable attributes, including texture and other visual details to improve authenticity. The generated seal is saved with a transparent background, allowing it to be seamlessly overlaid onto the document image for realistic integration.

To address the second challenge, we fully leverage the prior knowledge of pretrained text-conditioned image generative models, such as the Stable Diffusion Model. \textcolor{black}{The general architecture of the pretrained Stable Diffusion Model employed in our work, including its core Denoising U-Net component, is illustrated in Fig.~\ref{fig:unet}.} Based on this, we propose Seal2Real, an unsupervised method utilizing prompt prior learning (see Fig.~\ref{fig:prior_learning}) for the task of seal forgery using a diffusion model.

\begin{figure*}[!t]
  
  \begin{minipage}[b]{0.69\linewidth}  %
    \centering
    \includegraphics[width=1\linewidth]{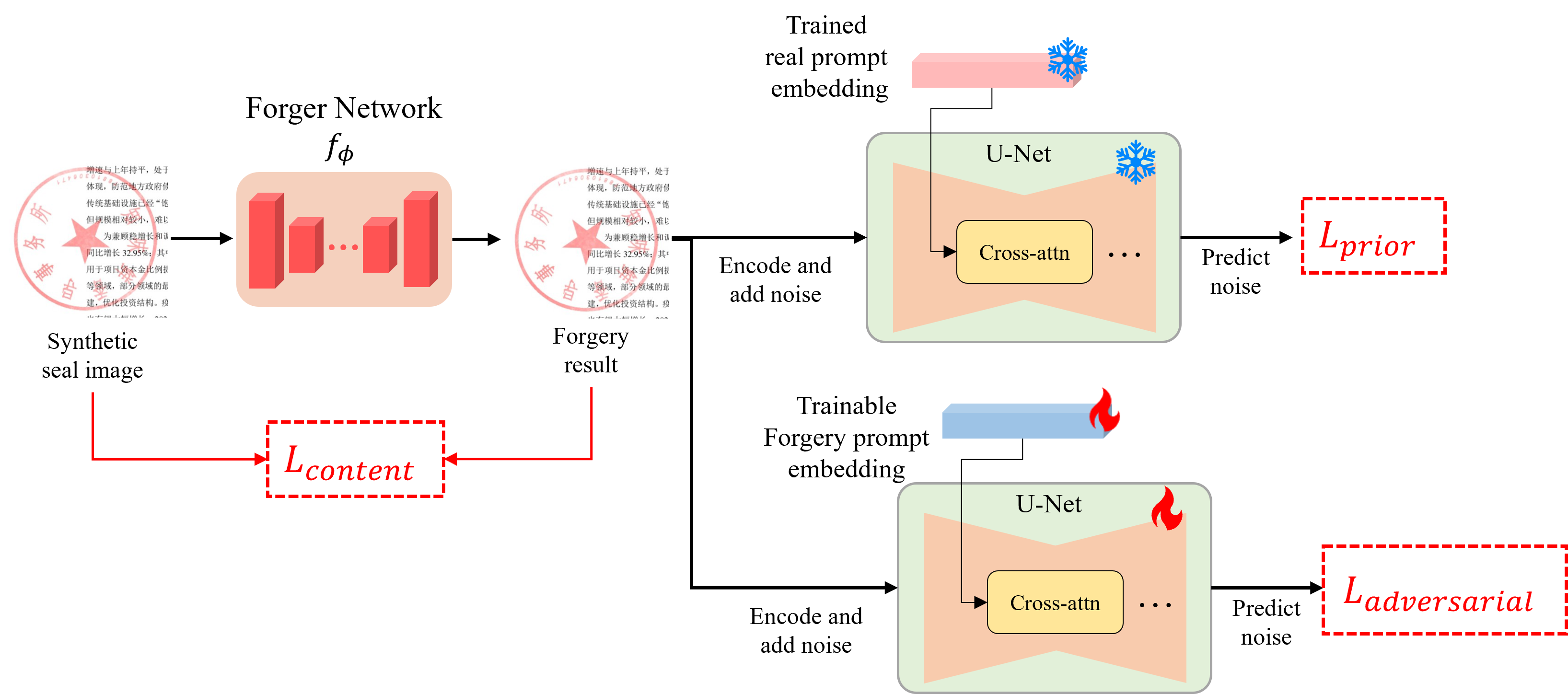}
    \centerline{(a)}\medskip
  \end{minipage}
  \hfill
  \begin{minipage}[b]{0.29\linewidth}  %
    \centering
    \includegraphics[width=1\linewidth]{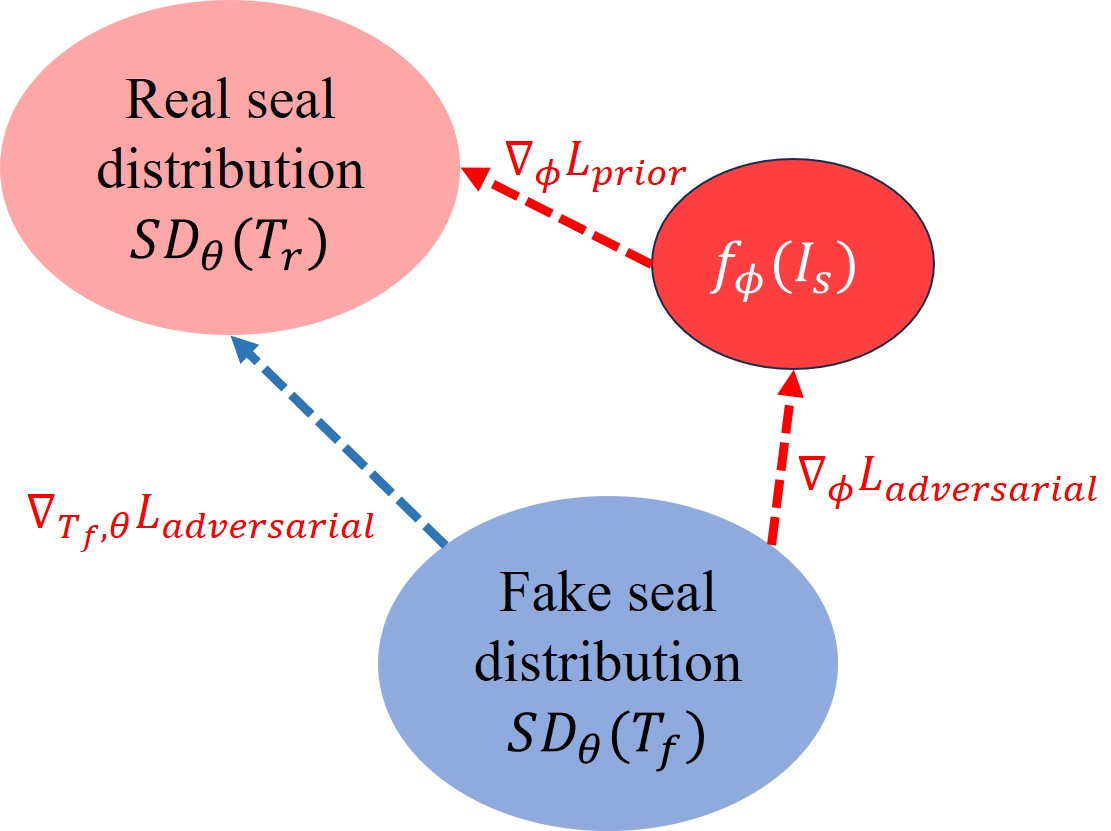}
    \centerline{(b)}\medskip
  \end{minipage}
  \caption{Illustration of the second stage learning framework. (a) \textbf{Forger network learning stage}. After the prompt learning stage, we use the prompt priors learned in the first stage to train our seal forger network. (b) Illustration of the modeled distribution shift in the second stage.}
  \label{fig:iterative_prompt_learning_framework}
\end{figure*}

The training of Seal2Real consists of two stages, as illustrated in Fig.~\ref{fig:iterative_prompt_learning_framework}. The first stage is the prompt learning stage, where prompts are learned and the diffusion model is fine-tuned to capture the distribution of both real and fake seal document images. The second stage is the forger learning stage, in which a forger network is trained to produce forged seal images with high realism.

\begin{figure}[t]

\begin{center}
          \includegraphics[width=0.75\linewidth]{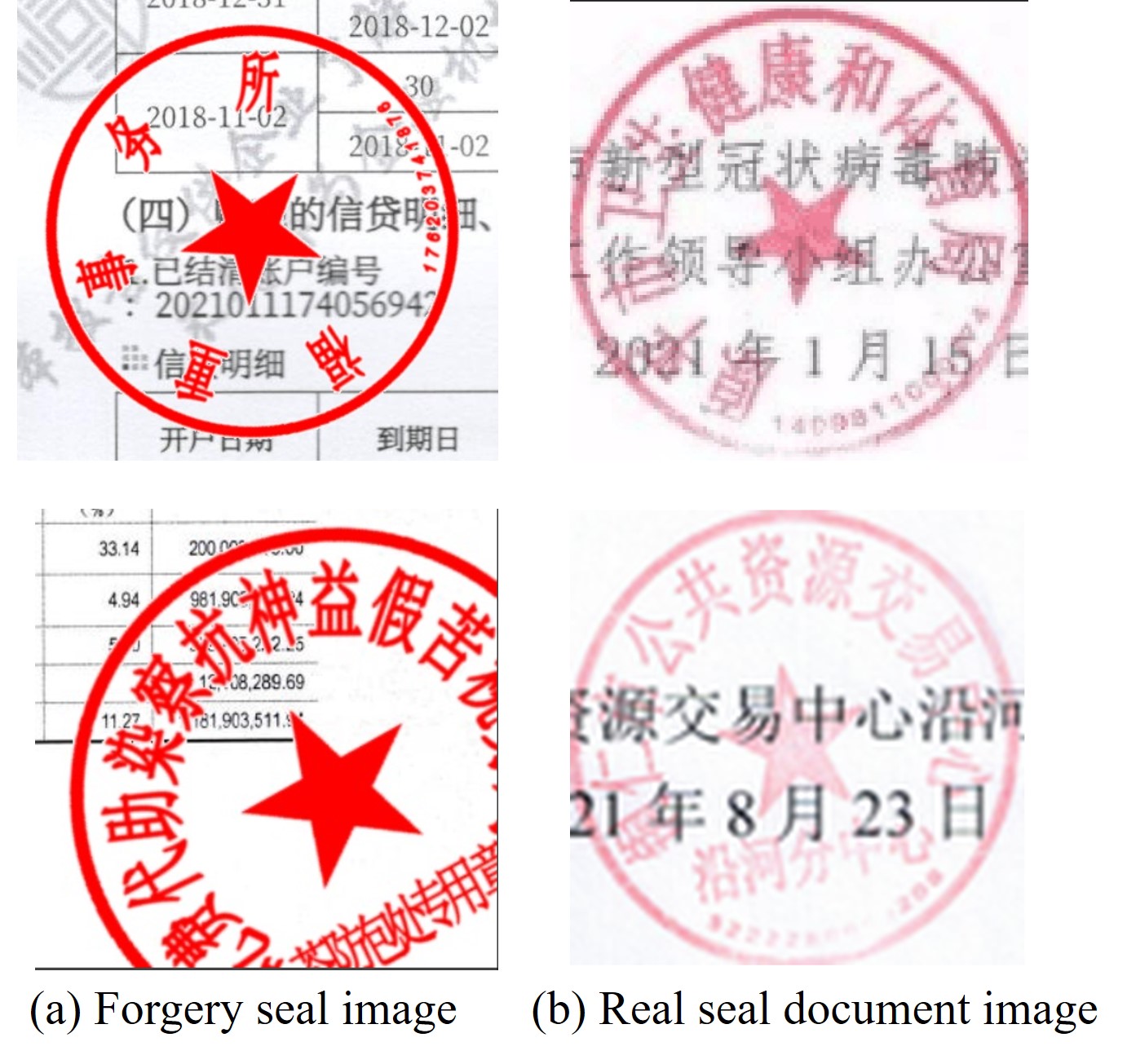}
    \end{center}
\caption{Illustration of our Seal-DB, consisting of a forgery part and a real part. The forgery seal images, also referred to as synthetic seal images, are generated in large quantities with high realism. These images are accompanied by paired labels, including segmentation masks (for seal segmentation), seal text (for text recognition under seals), and non-seal images (for seal removal). In contrast, the real part consists of unpaired images without annotations.}
\label{dataset}
\end{figure}

\subsection{Prior Learning Stage}\label{subsec:initial_prompt}

In the first stage, we aim to learn two distinct prompts: a real prompt, referring to real seal images, and a forgery (or fake) prompt, referring to all synthetic seal images. These prompts are learned using the pretrained Stable Diffusion Model and its simple loss function~\cite{RN3}, applied to our proposed Seal-DB dataset. As shown in Fig.~\ref{dataset}, Seal-DB comprises two parts: synthetic seal and real seal document images. The synthetic seal images are generated using our traditional seal synthesis method described in Sec.~\ref{sec:intro}.

The initial real and forgery prompts serve as coarse textual descriptions of real and fake seal images, as illustrated in Fig.~\ref{fig:iterative_prompt_learning_framework}(a). Specifically, we initialize the real prompt as ``A photo of document with real seal,'' with its corresponding embedding $T_{r} \in \mathbb{R}^{N\times 768}$, and the forgery prompt as ``A photo of document with fake seal,'' with embedding $T_{f} \in \mathbb{R}^{N\times 768}$~\cite{li2022clip}, where $N$ denotes the length of each prompt.

Given a real seal image $I_{r}$, we input $I_{r}$ and the real prompt $T_{r}$ into the pretrained Stable Diffusion Model $SD_{\theta}(I_r,t_r,T_r)$. The image $I_{r}$ is first encoded into a latent representation $z_r$, to which random Gaussian noise $\epsilon_r$ is added, resulting in the noised latent $z_{r,t_r}$. The model $SD_{\theta}(I_r,t_r,T_r)$ then estimates the added noise $\hat{\epsilon}_r$, where $\theta$ represents the parameters of the U-Net~\cite{ronneberger2015unet} in the Stable Diffusion Model~\cite{rombach2022high}. Similarly, for a synthetic seal image $I_{s}$, we input $I_{s}$ and the forgery prompt $T_{f}$ into $SD_{\theta}(I_s,t_f,T_f)$.

It is important to note that the noise terms $\epsilon_r, \epsilon_f$ and time steps $t_r$, $t_f$ are randomly sampled at each iteration to approximate the law of large numbers. Based on the text-conditioned noise estimation loss~\cite{miyake2025negative} in latent space, we apply the simple loss $\mathcal{L}_{prompts}$ to learn the real and forgery prompt pair while keeping the U-Net parameters $\theta$ frozen.
\begin{equation}
    \begin{aligned}
    \mathcal{L}_{prompts} = \mathbb{E}_{t_r,\epsilon_r} \min_{T_r} |SD_{\theta}(I_r,t_r,T_r) - \epsilon_r|^2 \\
    + \mathbb{E}_{t_f,\epsilon_f} \min_{T_f} |SD_{\theta}(I_s,t_f,T_f) - \epsilon_f|^2,
    \label{eq:prompts}
    \end{aligned}
\end{equation}
\textcolor{black}{where the output of $SD_{\theta}(I_r,t_r,T_r)$ is $\hat{\epsilon}_r$ \textcolor{black}{(the predicted noise for the real image case)}, and similarly $\hat{\epsilon}_f$ (not explicitly written) is the output $SD_{\theta}(I_s,t_f,T_f)$ (the predicted noise for the synthetic image case). Here, $\mathcal{L}_{prompts}$ represents the loss function specifically for optimizing the prompt embeddings $T_r$ and $T_f$. $\mathbb{E}_{t_r,\epsilon_r}$ denotes the expectation taken over randomly sampled diffusion timesteps $t_r$ and corresponding Gaussian noise $\epsilon_r$ (similarly for $t_f, \epsilon_f$). The $\min_{T_r}$ (and $\min_{T_f}$) indicates that the loss minimizes the subsequent term with respect to the learnable real prompt embedding $T_r$ (and forgery prompt $T_f$). $SD_{\theta}$ is the noise prediction function of the Stable Diffusion model with frozen parameters $\theta$. $I_r$ and $I_s$ are the input real and synthetic seal images, respectively. $t_r$ and $t_f$ are the randomly sampled timesteps. $\epsilon_r$ and $\epsilon_f$ are the ground truth Gaussian noise vectors added to the latent representations of $I_r$ and $I_s$ at their respective timesteps. The term $|\cdot|^2$ represents the squared L2 norm, calculating the squared Euclidean distance between the predicted noise ($\hat{\epsilon}_r$ or $\hat{\epsilon}_f$) and the actual added noise ($\epsilon_r$ or $\epsilon_f$).}

In addition, we fine-tune the U-Net parameters $\theta$ to better learn the distribution of real and fake seal images while keeping the prompts fixed:

\begin{equation}
    \begin{aligned}
    \mathcal{L}_{UNet} = \mathbb{E}_{t_r,\epsilon_r} \min_{\theta} |SD_{\theta}(I_r,t_r,T_r) - \epsilon_r|^2 \\
    + \mathbb{E}_{t_f,\epsilon_f} \min_{\theta} |SD_{\theta}(I_s,t_f,T_f) - \epsilon_f|^2,
    \label{eq:UNet}
    \end{aligned}
\end{equation}
\textcolor{black}{where $\mathcal{L}_{UNet}$ represents the loss function used for fine-tuning the U-Net parameters $\theta$ of the Stable Diffusion model ($SD_{\theta}$). The expectation terms $\mathbb{E}_{t_r,\epsilon_r}$ and $\mathbb{E}_{t_f,\epsilon_f}$, input images $I_r, I_s$, timesteps $t_r, t_f$, and ground truth noise vectors $\epsilon_r, \epsilon_f$ are defined as in Eq \ref{eq:prompts}. The key difference is that the optimization $\min_{\theta}$ is performed over the model parameters $\theta$, while the prompt embeddings $T_r$ and $T_f$ are held constant during this step. The objective remains to minimize the squared L2 distance ($|\cdot|^2$) between the noise predicted by the model ($SD_{\theta}$) and the actual added noise ($\epsilon_r$ or $\epsilon_f$).}

We alternate between prompt learning and U-Net fine-tuning until the generated results reach a visually satisfactory quality.

\subsection{Forger Network Learning Stage}

After learning the real and forgery prompts in the first stage, we proceed to train a forger network using a prompt-prior loss. As a baseline, we adopt a simple U-Net architecture~\cite{ronneberger2015unet} for our forger $f_{\phi}$ to enhance the realism of synthetic seal images and to validate the effectiveness of our framework.

Given a synthetic seal image $I_s$, the forger network $f_\phi$ produces a forged output $I_f = f_\phi(I_s)$, where the position, size, and textual content of the seal in $I_f$ are expected to remain consistent with those in $I_s$.

To train the forger network $f_\phi$, we introduce two loss functions: the prompt-prior loss $\mathcal{L}_{prior}$ and a content loss $\mathcal{L}_{content}$.

The prompt-prior loss leverages the learned prompt priors from the first stage to guide the training of the forger network:

\begin{equation}
    \begin{aligned}
    \mathcal{L}_{prior} = \mathbb{E}_{t_r,\epsilon_r} \min_{f_\phi} |SD_{\theta}(f_\phi(I_s), t_r, T_r) - \epsilon_r|^2,
    \end{aligned}
    \label{eq:prior}
\end{equation}
\textcolor{black}{where $\mathcal{L}_{prior}$ guides the forger network $f_\phi$ (optimized via $\min_{f_\phi}$) to produce images $f_\phi(I_s)$ that the frozen Stable Diffusion model $SD_{\theta}$, when conditioned on the learned real prompt $T_r$ and a random timestep $t_r$, perceives as having low noise $\epsilon_r$. The expectation $\mathbb{E}_{t_r,\epsilon_r}$ is over random timesteps and corresponding ground truth noise, $I_s$ is the input synthetic image, and $|\cdot|^2$ is the squared L2 norm. This encourages the forger's output to align with the distribution represented by the learned real prompt.}

The content loss encourages the forged result $I_f$ to preserve the same content and structural layout as the original synthetic seal document image $I_s$:

\begin{equation}
\label{eq:content_loss}
\mathcal{L}_{content} = \sum_{l=0}^{4} \alpha_l \cdot \|\Phi_{image}^{l}(I_{f}) - \Phi_{image}^{l}(I_{s})\|_2,
\end{equation}
\textcolor{black}{where $\alpha_l$ denotes the weight assigned to the $l$-th layer of the VGG network $\Phi$. Specifically, $\mathcal{L}_{content}$ is a perceptual loss calculated as a weighted sum over different layers $l$ (from 0 to 4). $\Phi_{image}^{l}(I_{f})$ represents the feature activation map extracted from the $l$-th layer of a pre-trained VGG network given the forged image $I_f = f_\phi(I_s)$, and $\Phi_{image}^{l}(I_{s})$ is the corresponding feature map for the original synthetic input $I_s$. $\|\cdot\|_2$ denotes the L2 norm measuring the difference between these feature maps, encouraging visual similarity at multiple semantic levels.}

The forger network is trained using a combined loss function that integrates both the prompt-prior and content losses:

\begin{equation}
\mathcal{L}_{forger} = \mathcal{L}_{prior} + w \cdot \mathcal{L}_{content},
\label{L-forger}
\end{equation}
\textcolor{black}{where $w$ is a balancing coefficient to control the relative contribution of the two loss terms. $\mathcal{L}_{forger}$ is the final objective function minimized to train the parameters $\phi$ of the forger network $f_\phi$. It linearly combines the prompt-prior loss $\mathcal{L}_{prior}$ (from Eq. \ref{eq:prior}) and the content loss $\mathcal{L}_{content}$ (from Eq. \ref{eq:content_loss}), with the hyperparameter $w$ controlling the influence of the content preservation term relative to the realism term guided by the learned prompt prior.}

While Eq.~\ref{eq:prior} plays a critical role in guiding the forger network to generate images aligned with the learned prompts-thereby enhancing the realism of the forgery-it may not sufficiently preserve the texture and structural details of the input image, potentially resulting in a completely different visual output. Conversely, Eq.~\ref{eq:content_loss} is essential for maintaining the content and structure of the original synthetic seal image. However, relying solely on this loss may fail to achieve a convincingly realistic appearance in the forged result.

To reconcile these objectives, Eq.~\ref{L-forger} introduces a composite loss function, $\mathcal{L}_{forger}$, which combines the prompt-prior loss and the content loss, weighted by a factor $w$. This integration is crucial to ensure that the output of the forger network not only appears realistic but also retains the essential visual characteristics of the input image. By balancing authenticity and fidelity, the forged image achieves both visual plausibility and structural consistency.
\begin{figure}[h]
 \begin{center}
    \includegraphics[width=1\linewidth]{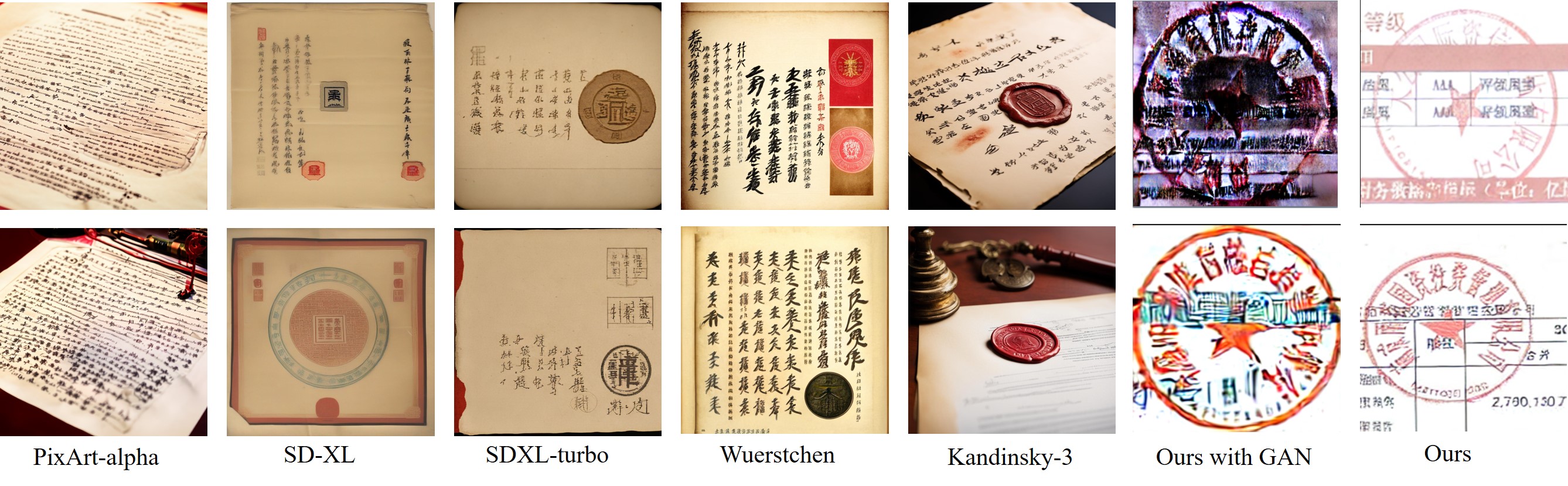}
    \end{center}
    
\caption{Comparison with several state-of-the-art diffusion-based text-to-image generation methods. Benefiting from the prior learning stage, our $SD_{\theta}$ is capable of generating high-quality real seal images that align well with the real seal distribution. The top and bottom rows present two visual samples generated by each method for comparison.}
\label{compare_sota}
\end{figure}

\begin{figure*}[t]
 \begin{center}
          \includegraphics[width=1\linewidth]{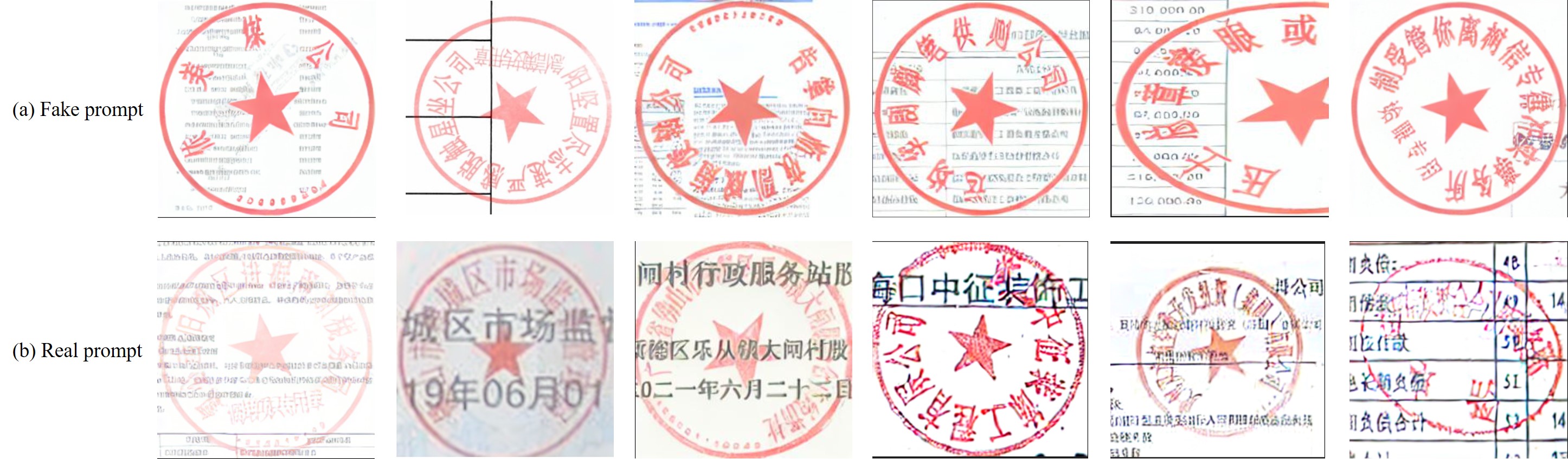}
    \end{center}

\caption{Our generated results after the prior learning stage: (a) Using the fake prompt in our $SD_{\theta}$ to generate a fake seal image from the fake seal distribution, and (b) Using the real prompt in our $SD_{\theta}$ to generate a real seal image from the real seal distribution. }
\label{compare}

\end{figure*}

We design the second stage as an alternating adversarial training process. First, we apply $\mathcal{L}_{content}$ to encourage the forged output to remain similar to $I_s$ in the pixel space. Then, we use $\mathcal{L}_{forger}$ to train the forger network $f_\phi$, enabling it to generate visually realistic forged seal images $I_f$. Once $f_\phi$ is sufficiently trained, we alternate by feeding $I_f$ into the Stable Diffusion model and fine-tune both the prompts and the U-Net for adversarial learning (as the saying goes, “the devil is one foot tall, virtue ten feet”):

\begin{equation}
    \begin{aligned}
    \mathcal{L}_{adversarial} = \mathbb{E}_{t_f,\epsilon_f} \min_{T_f,\theta} |SD_{\theta}(I_r, t_f, T_f) - \epsilon_r|^2 \\
    + \mathbb{E}_{t_f,\epsilon_f} \min_{\phi} |SD_{\theta}(f_\phi(I_s), t_f, T_f) - \epsilon_f|^2,
    \end{aligned}
    \label{eq:adversarial}
\end{equation}
\textcolor{black}{where $\mathcal{L}_{adversarial}$ represents the loss function for the adversarial training stage. The expectation $\mathbb{E}_{t_f,\epsilon_f}$ is taken over random timesteps $t_f$ and corresponding Gaussian noise $\epsilon_f$. The first term involves minimizing the squared L2 norm ($|\cdot|^2$) between the noise predicted by the Stable Diffusion model $SD_{\theta}$ for a real image $I_r$ (conditioned on the fake prompt $T_f$ and timestep $t_f$) and the ground truth noise $\epsilon_r$; this minimization is performed jointly over the fake prompt embedding $T_f$ and the SD model parameters $\theta$. The second term involves minimizing the squared L2 norm between the noise predicted by $SD_{\theta}$ for the forger's output $f_\phi(I_s)$ (conditioned on the fake prompt $T_f$ and timestep $t_f$) and the ground truth noise $\epsilon_f$; this minimization is performed over the parameters $\phi$ of the forger network $f_\phi$. Note that $I_s$ is the original synthetic input image.}

This alternating adversarial training continues until the generated results achieve satisfactory visual quality.

\subsection{Training with Frozen Encoder}
In our setup, the CLIP text encoder is kept frozen, meaning its parameters remain unchanged throughout training. This is essential for maintaining consistent text representations, which enables effective control over the generation of real and synthetic seal images based on textual prompts. During training, real prompts are paired with actual seal images, while fake prompts are associated with synthetic seal images. This dichotomous pairing ensures that the fine-tuned Stable Diffusion model learns to distinguish effectively between the two underlying distributions. As a result, by leveraging distinct textual prompts, our model is capable of generating outputs that accurately reflect the intended nature-real or synthetic-of the input prompt.

\section{Experiments}
\label{sec:pagestyle}
\subsection{Dataset and Setup}
We utilize the unsupervised document seal generation and realization method described above to construct our dataset, Seal-DB. Seal-DB consists of 10K synthetic seal images and 10K scanned or photographed images of real seals. All synthetic seal images are generated using our proposed method, with their texture and color distributions closely resembling those of real seal images. Furthermore, leveraging our trained model and the existing dataset, additional data can be selectively generated to further expand the dataset in a controlled and scalable manner.

\begin{figure}[!ht]
\begin{center}
    \includegraphics[width=0.85\linewidth]{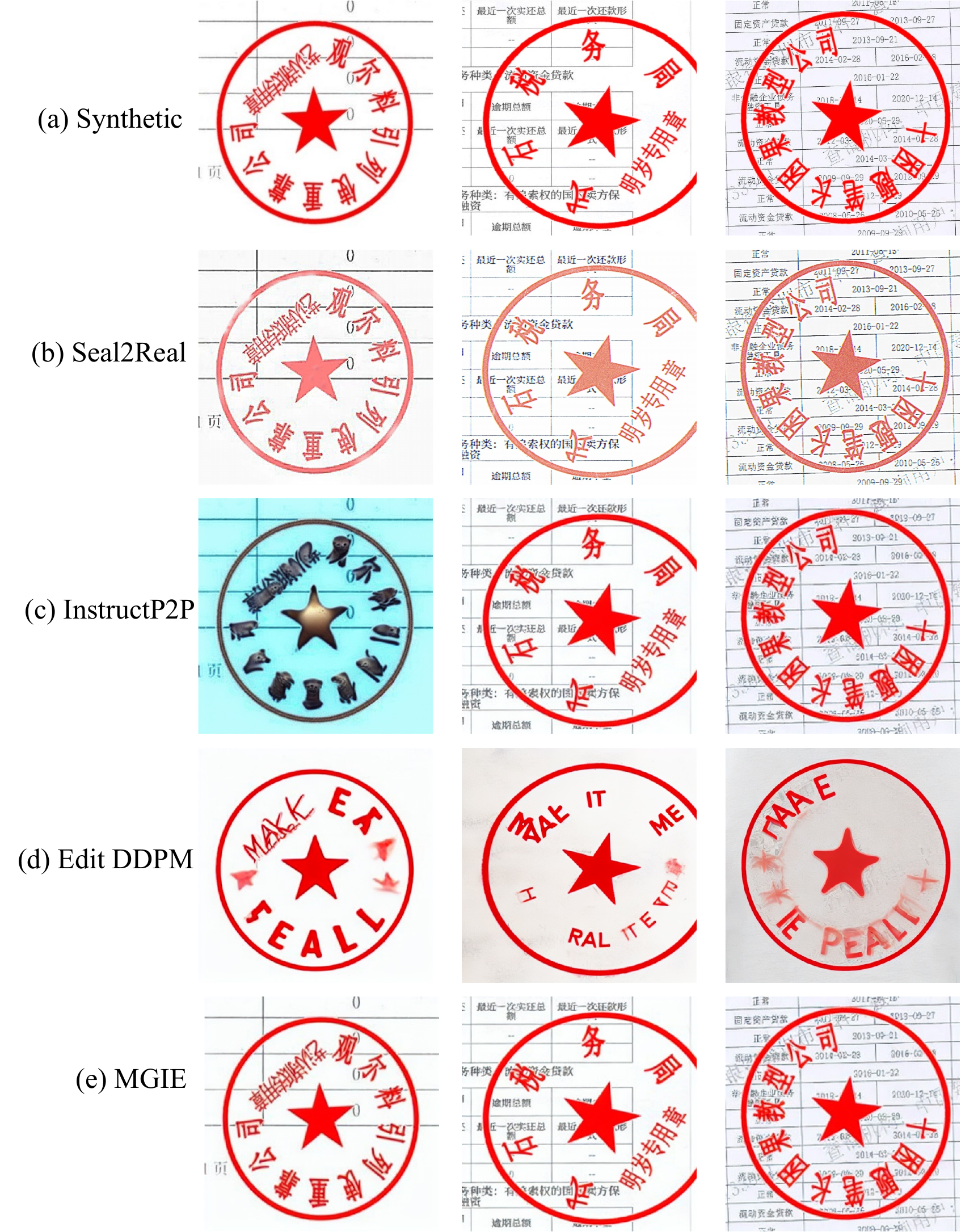}
    \end{center}
\caption{Comparison between the traditional synthetic seal image and our generative forgery result. (a) Synthetic seal image. (b) Forgery seal image generated by our Seal2Real. (c)–(e) Edited results produced by several state-of-the-art text-conditioned image editing methods. 
Our method (b) demonstrates a significant improvement in texture realism and overall authenticity, with the generated seals exhibiting both accurate structural form and a lifelike visual appearance.
}
\label{forgery result}
\end{figure}

To train our method and validate its effectiveness, we conduct experiments on our self-constructed Seal-DB dataset. We evaluate performance on downstream tasks, including seal segmentation and text recognition under seals. All experiments are carried out using six NVIDIA A5000 GPUs. Throughout the training process, we employ a pre-trained Stable Diffusion model as the foundation.

The optimization is performed using the Adam algorithm with a learning rate of 2e-6. Additionally, we compare the performance across different training datasets to assess the impact of data quality and diversity. The experimental results demonstrate that our dataset significantly enhances the performance of downstream tasks, validating both the effectiveness of our generative framework and the utility of Seal-DB.

\subsection{Analysis of Experimental Results}
\textbf{Results}. Using our method, we generate images under both the ``fake prompt'' and ``real prompt'' conditions. Compared to traditional synthetic images, those generated under the ``fake prompt'' condition show a higher degree of similarity to real seal images, exhibiting more realistic color distribution, lighting, and texture. Fig.~\ref{compare} illustrates the generated results under both prompt conditions, highlighting the effectiveness of our approach.

To further demonstrate the necessity of our method, Fig.~\ref{compare_sota} presents results from several state-of-the-art text-to-image generation models, including PixArt-$\alpha$~\cite{chen2023pixart}, SD-XL, SDXL-turbo~\cite{sauer2023adversarial}, Wuerstechen~\cite{perniaswurstchen}, and Kandinsky-3~\cite{arkhipkin2023kandinsky}, along with an ablation study using a GAN-based baseline. These comparisons clearly illustrate the superiority and robustness of employing a pretrained diffusion model over training a GAN from scratch.

Additionally, we conducted a user study by distributing 100 surveys to collect human ratings on the quality of the generated images. The results of this evaluation are summarized in Tab.~\ref{tab:user_study}.

\begin{table}[t]
\centering
\begin{tabular}{c|c|c|c|c|c}
\hline
\multirow{2}{*}{\makecell{Image \\ ID}} & \multirow{2}{*}{Synthetic} & \multicolumn{2}{c|}{Prompt Type} & \multirow{2}{*}{\makecell{The Forger \\ Network}} & \multirow{2}{*}{GAN} \\
& & Fake & Real & & \\
\hline
1       & 5.26 & 6.35 & 7.89 & 6.12 & 6.10  \\
2       & 6.13 & 7.43 & 7.95 & 6.78 & 6.79  \\
3       & 5.63 & 6.89 & 7.98 & 6.26 & 6.12  \\
4       & 5.58 & 6.79 & 7.26 & 6.28 & 6.31  \\
5       & 5.70 & 6.88 & 7.58 & 5.99 & 6.02  \\
6       & 4.96 & 6.46 & 7.25 & 5.85 & 5.83  \\
7       & 5.13 & 5.96 & 6.98 & 5.36 & 5.39  \\
8       & 4.98 & 5.85 & 7.99 & 5.49 & 5.17  \\
9       & 5.65 & 6.78 & 8.02 & 5.97 & 5.99  \\
10      & 6.32 & 7.55 & 8.59 & 7.02 & 6.85  \\
\hline
average & 5.53 & 6.69 & 7.75 & 6.11 & 6.06 \\
\bottomrule
\end{tabular}
\caption{User study scores for 10 generated images.}

\label{tab:user_study}
\end{table}

\begin{figure}[!ht]
\begin{center}
    \includegraphics[width=0.95\linewidth]{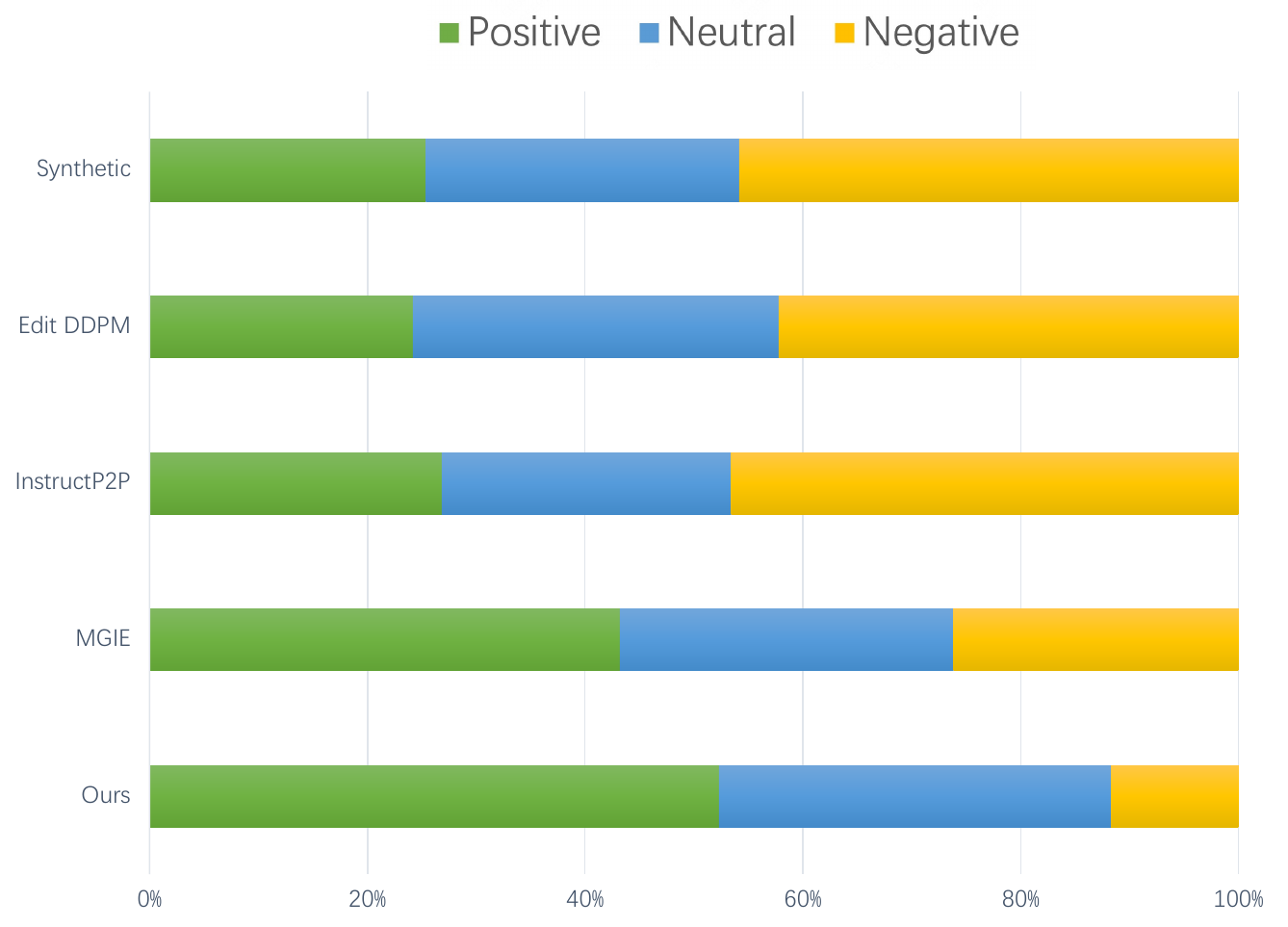}
    \end{center}
\caption{User study on the forgery results. A positive rating indicates that the user perceives the image as resembling a real seal; a neutral rating suggests uncertainty, and a negative rating implies that the user believes it does not resemble a real seal.
}
\label{user}
\end{figure}
\textbf{Impact of the forger network}. In the second stage of our training pipeline, we introduce a forger network designed to enhance the authenticity of generated seals and progressively align them with the characteristics of real seals. The integration of this forger network results in continuous improvements in the visual quality of the generated data, while also contributing to the effectiveness of training for downstream tasks.

Fig.~\ref{forgery result} illustrates the visual differences in synthetic images before and after applying the forger network. In addition, we compare our Seal2Real framework with several state-of-the-art text-conditioned image editing methods, as such methods might appear suitable for this task. Specifically, we evaluate InstructP2P~\cite{brooks2022instructpix2pix}, EditDDPM~\cite{huberman2023edit}, and MGIE~\cite{fu2023guiding}, using the editing instruction: ``Make this seal more like the real one."

Visual comparison results are presented in Fig.~\ref{forgery result}, and corresponding quantitative results from the user study are shown in Fig.~\ref{user}.

\textbf{Evaluation of downstream tasks}. To evaluate the effectiveness of our method on data quality and utility, we select three representative downstream tasks: seal segmentation, authentication of real versus fake seals, and text recognition under seals.

For each task, we adopt established baseline methods and modify only the training datasets to assess the impact of our synthesized data. Specifically, for seal segmentation, we use the U-Net architecture~\cite{ronneberger2015unet}; for authenticity classification, we employ a CLIP-based approach~\cite{radford2021learning}; and for OCR in documents containing seals, we utilize SwinTextSpotter~\cite{huang2022swintextspotter}. All evaluations are conducted using the same quantity of training data, comparing the original dataset with our generated dataset to highlight the improvements enabled by our method.

Experimental results demonstrate that our method produces data with a high degree of realism, significantly improving the performance of downstream tasks. These results are summarized in Tab.~\ref{tab:downstream}, supporting our earlier claim in Sec.~\ref{sec:intro} that seal-related tasks are heavily dependent on labeled data, and that generating realistic datasets is critical to their success.

All downstream tasks are trained in a fully supervised manner using the dataset generated by our forger network. The consistent improvements across all tasks validate the effectiveness of our forgery framework and underscore the importance of high-quality, realistic synthetic data in training pipelines. The enhanced results in seal segmentation, authenticity classification, and text recognition under seals confirm the value of our forged dataset, reinforcing our assertion that realistic data generation plays a pivotal role in advancing seal-related document analysis and applications.

\begin{table}[t]

\centering
\begin{tabular}{cccc}
\hline
Metrics  & Segmentation\quad\quad & Identification\quad\quad & Word Recognition\quad\quad   \\
\hline
 \multicolumn{4}{c}{Traditional Synthetic Seal Dataset}  \\
\hline
MIOU(\%)       & 78.3  & - & -   \\
Accuracy(\%)       & - & 85.63 & 75.34   \\
\hline
\multicolumn{4}{c}{Our Realisation Dataset}   \\
\hline
MIOU(\%)        &  91.5 & - & -   \\
Accuracy(\%)       & - & 90.23 & 81.59   \\
\bottomrule
\hline
\end{tabular}
\caption{Evaluation of three downstream tasks: seal segmentation in document images, authentication of real and fake seals, and word recognition under seal interference.}

\label{tab:downstream}
\end{table}

\section{Conclusion}
\label{sec13}
\label{sec:typestyle}

In this paper, we presented Seal2Real, a novel framework designed to address the critical challenge of data scarcity in document seal analysis. By employing a prompt prior learning architecture built upon a pre-trained Stable Diffusion model, Seal2Real effectively transfers powerful generative capabilities to the unsupervised domain of seal image synthesis. This allows for the creation of Seal-DB, a large-scale dataset of realistic synthetic seal images paired with automatic ground-truth labels. \textcolor{black}{Our experiments confirm that data generated by Seal2Real significantly enhances the performance of various downstream seal-related tasks compared to traditional synthesis methods, demonstrating the value of high-fidelity synthetic data.} \textcolor{black}{While the approach is promising, its performance is influenced by the diversity of the initial real seal data used for learning priors, indicating clear paths for future enhancement.} Overall, Seal2Real offers a valuable contribution towards advancing research and development in automated document processing involving seals.

\section{Discussion}\label{sec:discussion}
\textcolor{black}{The significant improvements in downstream task performance (Table 2) validate the effectiveness of the Seal2Real framework. We attribute this success primarily to the enhanced realism achieved through our prompt prior learning mechanism. By learning distinct representations for 'real' and 'fake' seals guided by comparing unlabeled real images and synthetic examples, the model generates data that better reflects real-world seal characteristics, leading to improved generalization for downstream models compared to traditional synthesis methods.}

\textcolor{black}{Despite these positive results, Seal2Real's performance is inherently linked to the initial set of real, unlabeled seal images (currently 10K) used for learning priors. The scale and diversity of this dataset directly impact the learned representations; potential biases within this data (e.g., prevalence of certain seal types) could limit the diversity of generated outputs and risk overfitting the priors. Accurately rendering extremely fine details or simulating complex degradation patterns also remains an area for improvement.}

\textcolor{black}{Future work should focus on mitigating these limitations. Incorporating larger and, crucially, more diverse real seal collections for prior learning is a key priority. We also plan to explore techniques for more explicit generation control, such as style conditioning, and investigate the potential of semi-supervised or few-shot learning approaches to enhance realism and diversity, especially when real data is limited. Finally, adapting the core concept of learning priors from synthetic/real pairs to other document elements like signatures or stamps presents a promising direction for extending the applicability of this framework.}

\section*{Acknowledgements}
This work was supported by the Shenzhen Science and Technology Program (JSGG20220831105002004) and Shenzhen Key Laboratory of Computer Vision and Pattern Recognition.

\section*{Declarations}
\textbf{Conflict of interest.} The authors declare that there are no conflict of
interests. All data included in this study are available upon request by
contact with the corresponding author.

 \bibliographystyle{elsarticle-num} 
 \bibliography{main}

\end{document}